\documentclass{article}
\usepackage{ijcai24}

\usepackage{times}
\usepackage{soul}
\usepackage{url}
\usepackage[hidelinks]{hyperref}
\usepackage[utf8]{inputenc}
\usepackage[small]{caption}
\usepackage{graphicx}
\usepackage{amsmath}
\usepackage{amsthm}
\usepackage{booktabs}
\usepackage{algorithm}
\usepackage[switch]{lineno}
\usepackage{multirow}

\usepackage{amsmath} 
\usepackage{algorithm}
\usepackage{amssymb}
\usepackage[noend]{algpseudocode}
\usepackage{graphicx}  
\usepackage{stfloats}   
\usepackage{subfigure}  
\usepackage{caption}   
\usepackage{setspace}  
\usepackage{color}
\usepackage{lipsum}

\title{\includegraphics[width=30pt]{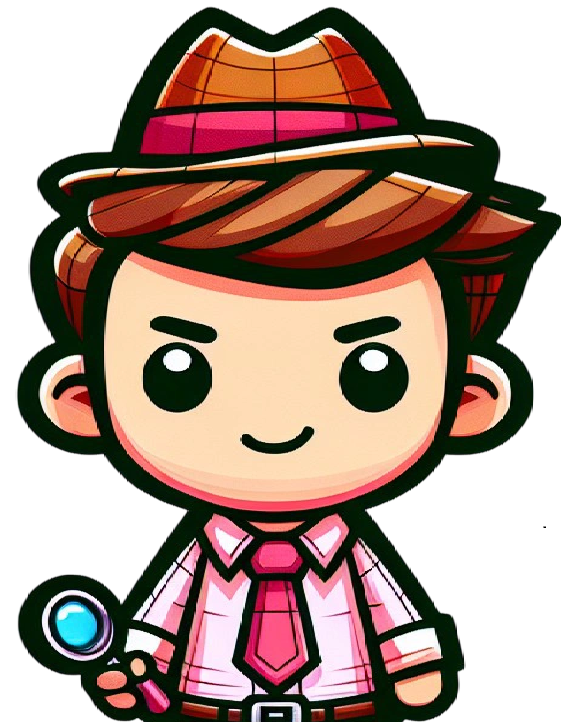}TextSleuth: Towards Explainable Tampered Text Detection}

\author{
    Chenfan Qu$^{1,2}$, Jian Liu$^2$, Haoxing Chen$^2$, Baihan Yu$^2$,\\ Jingjing Liu$^2$, Weiqiang Wang$^2$, Lianwen Jin$^1$ 
    \affiliations
    $^1$South China University of Technology
    $^2$Ant Group
    \emails
    202221012612@mail.scut.edu.cn, rex.lj@antgroup.com, eelwjin@scut.edu.cn
}

\begin{document}

\maketitle

\begin{abstract}
    Recently, tampered text detection has attracted increasing attention due to its essential role in information security. Although existing methods can detect the tampered text region, the interpretation of such detection remains unclear, making the prediction unreliable. To address this problem, we propose to explain the basis of tampered text detection with natural language via large multimodal models. To fill the data gap for this task, we propose a large-scale, comprehensive dataset, ETTD, which contains both pixel-level annotations for tampered text region and natural language annotations describing the anomaly of the tampered text. Multiple methods are employed to improve the quality of the proposed data. For example, elaborate queries are introduced to generate high-quality anomaly descriptions with GPT4o. A fused mask prompt is proposed to reduce confusion when querying GPT4o to generate anomaly descriptions. To automatically filter out low-quality annotations, we also propose to prompt GPT4o to recognize tampered texts before describing the anomaly, and to filter out the responses with low OCR accuracy. To further improve explainable tampered text detection, we propose a \textbf{simple yet effective} model called TextSleuth, which achieves improved fine-grained perception and cross-domain generalization by focusing on the suspected region, with a two-stage analysis paradigm and an auxiliary grounding prompt. Extensive experiments on both the ETTD dataset and the public dataset have verified the effectiveness of the proposed methods. In-depth analysis is also provided to inspire further research. Our dataset and code will be open-source.
\end{abstract}

\section{Introduction}

Text image is one of the most important information carriers in today's society, containing a large amount of sensitive and private information~\cite{chen2024cma}. With the rapid development of image processing technologies, sensitive text information can be more easily manipulated for malicious purposes, such as fraud, posing serious risks to information security~\cite{dong2024robust}. Consequently, tampered text detection has become a major research topic in recent years~\cite{qu2024omniimlunifiedimagemanipulation}. It is crucial to develop effective and reliable methods for detecting tampered text images.

\begin{figure}
 \centering
 \includegraphics[width=0.5\textwidth]{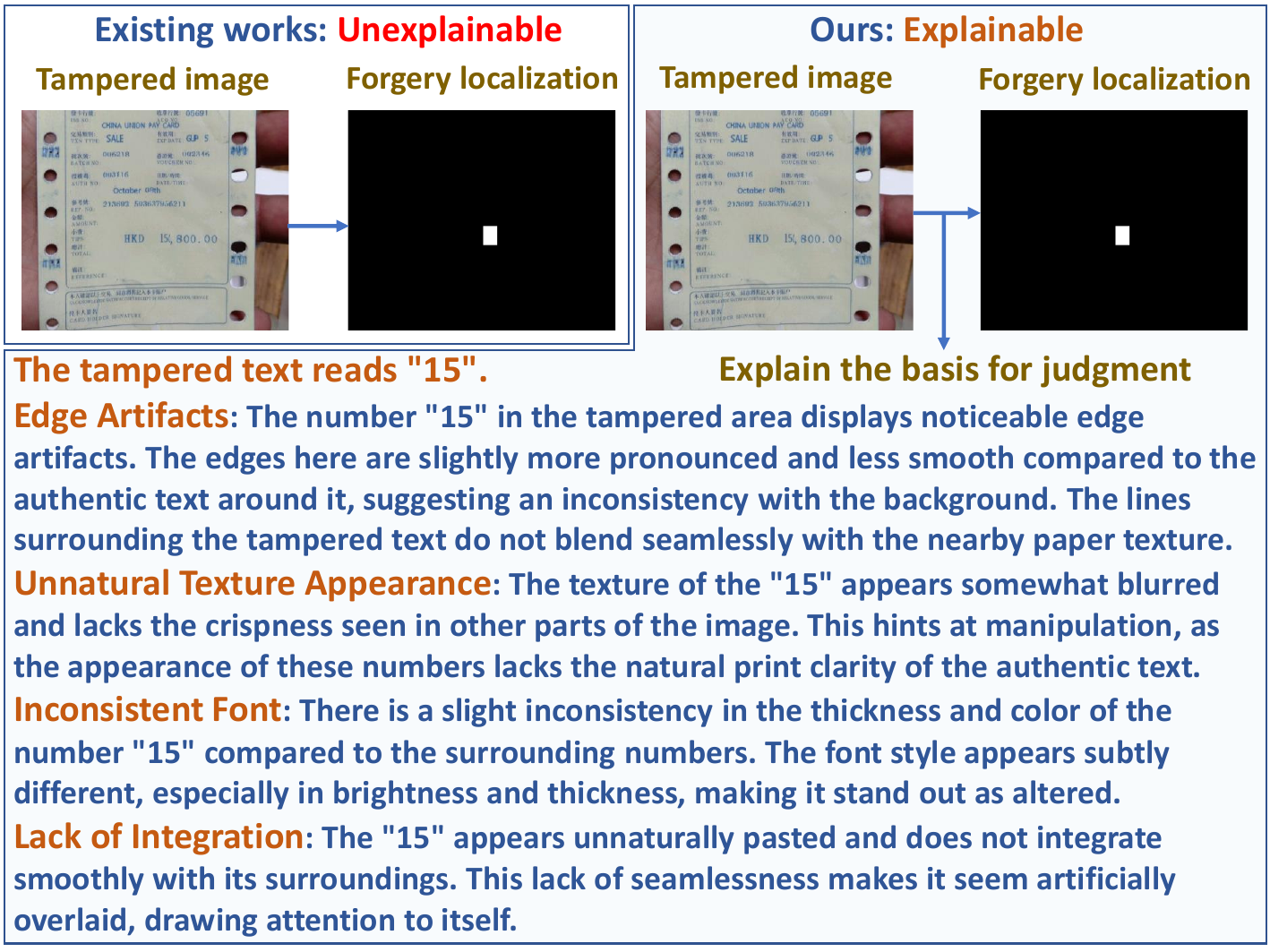}
 \caption{We propose to both detect the tampered text region and explain the basis for the detection in natural language, making the prediction more reliable. We construct the first dataset and propose a novel model for the explainable tampered text detection task.}
 \label{fig: teaser}
 \vspace{-0.3cm}
 \end{figure}

Existing works model tampered text detection as semantic segmentation~\cite{shao2023progressive} or object detection~\cite{qu2024generalizedtamperedscenetext}, with the aim of interpreting the basis for image forgery classification by predicting tampered regions. Despite the progress made in recent years, such fine-grained predictions are still black-box and cannot provide a convincing basis to support the judgement, leading to unreliable results.  

To provide more reliable predictions for tampered text detection, we propose to leverage multimodal large models to both detect tampered text regions and explain the basis for their detection in natural language. Given the absence of dataset for interpretable tampered text detection, we construct the Explainable Tampered Text Detection (ETTD) dataset. To ensure the comprehensiveness of the data, we collect multilingual card images, document images and scene text images from the Internet and the existing text-rich datasets such as ICDAR2017~\cite{nayef2017icdar2017} and LSVT~\cite{lsvt}. And we perform text tampering on the collected data with various methods, including traditional methods copy-move, splicing, and the deep generative method DiffUTE~\cite{chen2024diffute}. Poisson Blending~\cite{perez2023poisson} is employed to reduce the visual inconsistency around tampered region. Finally, we 12,000 tampered text images with accurate pixel-level annotations of the tampered region and 10,500 authentic text images. The large-scale of our data notably alleviates the data hunger of deep models. The images are split to three parts, ETTD-Train, ETTD-Test and ETTD-CD, the two test sets have the same and different distributions with the ETTD-Train respectively, allowing both in-domain and cross-domain evaluation.

With the obtained tampered text, we utilize GPT4o to generate the description of both visual and linguistic anomalies caused by text tampering, and to generate the text recognition result for specifying the target tampered text. To achieve this, we prompt the GPT4o with a novel elaborate query, the tampered image and its corresponding mask annotation indicating the tampered region. However, since text is mostly dense and has similar location and shape, directly inputting the binary mask, as existing work~\cite{xu2024fakeshield} does will cause severe confusion to the GPT4o, making it unclear which is the actual tampered text. To solve this problem, we propose to fuse the binary mask into the original tampered image with pixel-wise weighting. With the proposed fused mask prompt, the GPT4o has a much better understanding of the location of the target region, which in turn significantly reduces the errors and obviously improves the annotation quality. In addition, the GPT4o's output is not always correct and manual verification is costly. Inspired by the fact that incorrect detection of manipulated text leads to unclear perception and poor anomaly description, we further propose to address this issue by automatically filtering out the annotation based on the accuracy of the manipulated text OCR.

The tiny area and visual consistency of tampered text~\cite{wang2022tic13} pose multiple challenges for explainable tampered text detection, making it difficult for existing methods to achieve good enough performance. For example, misidentification of tampered text leads to incorrect anomaly description, difficulty in finding tampered text weakens the analysis quality, and increases the risk of overfitting to unrelated background styles. To this end, we propose a novel \textbf{simple-yet-effective} model termed as TextSleuth. Specifically, an extra RCNN~\cite{ren2015fasterrcnn} based text detection module initially scans the image and predicts the location of the tampered text with cascaded RoI heads. The initial prediction of tampered region is converted into a grounding prompt and fed into the large language model along with the image tokens and the original question to obtain the final prediction. The proposed two-stage analysis paradigm and auxiliary prompt in TextSleuth effectively minimizes errors, improves explanation quality and cross-domain generalization by drawing the model's special attention to the anomaly region and helping it to learn more general features. In addition, since the reference grounding comprehension task is mostly involved in the pre-training stage of large models~\cite{internvl2}, the proposed auxiliary grounding prompt can reduce comprehension difficulty and alleviate forgetting.

Both our proposed ETTD dataset and TextSleuth model are the first efforts in the field of interpretable tampered text detection. Extensive experiments have confirmed that the proposed TextSleuth significantly improves upon the baseline model, outperforming existing methods by a large margin on both the proposed ETTD dataset and the public Tampered IC-13~\cite{wang2022tic13} dataset, demonstrating strong in-domain and cross-domain generalization capabilities. In-depth analysis is also provided to inspire further work in the field of interpretable tampered text detection.

In summary, the our main contribution is fourfold:
\begin{itemize}
    \item We propose a novel \textcolor{blue}{\textbf{task}}, explainable tampered text detection, which aims to provide reliable prediction by describing the anomalies of tampered text in natural language, serving as a pioneering effort in this field.
    \item We obtain the data annotation for this task by prompting GPT4o with elaborate queries. We propose effective methods to improve the quality of the annotations.~For example, a fused mask prompt to reduce model confusion and a novel method to automatically filter out bad responses. Based on these, we construct the ETTD dataset, which is \textcolor{blue}{\textbf{the first}}, large-scale and comprehensive \textcolor{blue}{\textbf{dataset}} for explainable tampered text detection.   
    \item We propose \textcolor{blue}{\textbf{the first}} multimodal large \textcolor{blue}{\textbf{model}} TextSleuth for interpretable tampered text detection, which achieves state-of-the-art performance with a two-stage analysis paradigm and a novel auxiliary prompt.
    \item Extensive experiments are conducted. Valuable conclusions and insights are provided through in-depth analysis, inspiring the further research in this field.
\end{itemize}

\section{Related works}

\subsection{Tampered Text Detection}
Early work on tampered text detection is achieved by printer classification~\cite{lampert2006printing} or template matching~\cite{ahmed2014forgery}, which is only applicable to scanned documents and does not work well for photographed documents~\cite{dong2024robust}. DocTamper~\cite{CVPR2023DocTamper} is the first large-scale comprehensive dataset for tampered text detection in arbitrary style documents. It is created by a novel method Selective Tampering Generation~\cite{CVPR2023DocTamper}, which effectively reduces the cost of high-quality training data. DTD~\cite{CVPR2023DocTamper} is proposed to detect visually consistent tampering in documents through examining the continuity of the block artifacts grids. The SACP~\cite{sacp} and RTM~\cite{luo2024RTM} datasets provide large-scale manually forged documents. Tampered IC13~\cite{wang2022tic13} dataset is the first work on tampered scene text detection, containing about 300 scene text images tampered by SR-Net~\cite{ACMMM_SRNet}. The OSFT dataset~\cite{qu2024generalizedtamperedscenetext} significantly improves Tampered IC-13 with realistic forgeries created by nine latest deep generative models. CAFTB-Net~\cite{song2024cross} benefits from noise-domain modeling and cross-attention mechanism. DTL~\cite{shao2025delving} improves model robustness with latent manifold adversarial training. Omni-IML~\cite{qu2024omniimlunifiedimagemanipulation} unifies image forensics across different text image types. Despite the progress made in recent years, existing work on tampered text detection can still only localize the tampered region in an unreliable black-box manner, unable to explain the basis of its judgement in natural language.

\begin{table*}[]
\setlength{\tabcolsep}{2pt}
\caption{A brief summary of the ETTD dataset statistics. "Forged Area" denotes the area ratio of tampered text.}
\vspace{-0.2cm}
\begin{tabular}{ccccccccccccc}
\hline
Dataset &  & Image types &  & Image source &  & Languages &  & Tampering methods \& Numbers &  & Authentic num. & \multicolumn{1}{l}{} & \multicolumn{1}{l}{Forged Area} \\ \cline{1-1} \cline{3-3} \cline{5-5} \cline{7-7} \cline{9-9} \cline{11-11} \cline{13-13} 
\multirow{2}{*}{ETTD-Train} &  & \multirow{4}{*}{\begin{tabular}[c]{@{}c@{}}Documents,\\ ID cards,\\ scene texts,\\ etc.\end{tabular}} &  & \multirow{4}{*}{\begin{tabular}[c]{@{}c@{}}Internet,\\ ICDAR2013,\\ ICDAR2017,\\ LSVT\end{tabular}} &  & \multirow{2}{*}{EN, CH} &  & \multirow{2}{*}{\begin{tabular}[c]{@{}c@{}}Total (10400): DiffUTE (800),\\ CopyMove (4800), Splicing (4800)\end{tabular}} &  & \multirow{2}{*}{9600} &  & \multirow{2}{*}{0.0268} \\
 &  &  &  &  &  &  &  &  &  &  &  &  \\
\multirow{2}{*}{ETTD-Test} &  &  &  &  &  & \multirow{2}{*}{EN, CH} &  & \multirow{2}{*}{\begin{tabular}[c]{@{}c@{}}Total (600): DiffUTE (200),\\ CopyMove (200), Splicing (200)\end{tabular}} &  & \multirow{2}{*}{400} &  & \multirow{2}{*}{0.0202} \\
 &  &  &  &  &  &  &  &  &  &  &  &  \\
\multirow{2}{*}{ETTD-CD} &  & \multirow{2}{*}{scene text} &  & \multirow{2}{*}{ICDAR2013} &  & \multirow{2}{*}{EN} &  & \multirow{2}{*}{\begin{tabular}[c]{@{}c@{}}Total (1000): Copymove (500),\\ Splicing (500)\end{tabular}} &  & \multirow{2}{*}{500} &  & \multirow{2}{*}{0.0608} \\
 &  &  &  &  &  &  &  &  &  &  & \multicolumn{1}{l}{} &  \\ \hline
\end{tabular}
\label{tab: data}
\vspace{-0.2cm}
\end{table*}

\begin{figure*}[t!]
    \centering
    \setlength{\abovecaptionskip}{3pt}
    \includegraphics[width=1.0\linewidth]{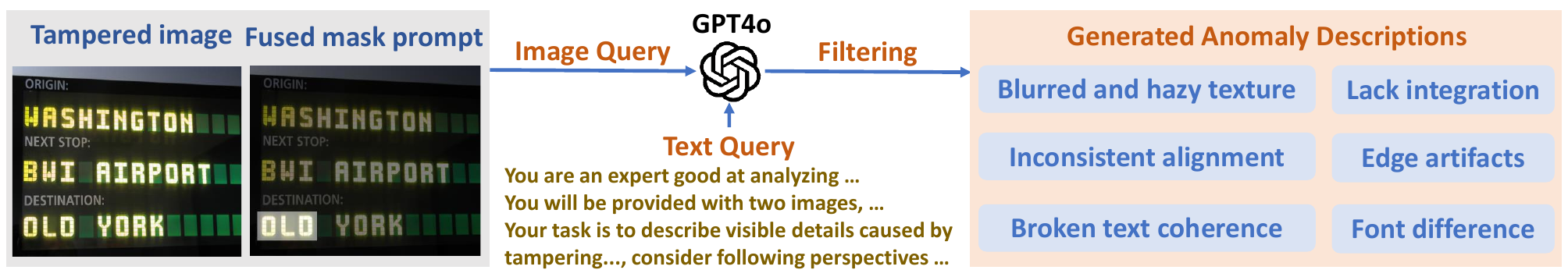}
    \caption{The pipeline for obtaining the textual anomaly description for the tampered text.}
    \label{fig: main-fig}
    \vspace{-0.4cm}
\end{figure*}

\subsection{Explainable Image Forgery Detection}
Recently, some works achieve explainable image forensics through multimodal large language models. FFAA~\cite{huang2024ffaa} utilizes GPT4o to generate detailed basis description for the identification judgment. MMTT~\cite{mmtt} further proposes to perform in-depth face forgery analysis by observing the facial organs one by one. FakeShield~\cite{xu2024fakeshield} leverages GPT4o to create anomaly description for natural style image forgery and introduces a new model based on LISA~\cite{lai2024lisa}. EditScout~\cite{nguyen2024editscout} focuses on the detection of image forgeries that generated by diffusion models and also develops a new model based on LISA. ForgeryGPT~\cite{li2024forgerygpt} improves interpretable natural image forensics with binary mask prompt. FakeBench~\cite{li2024fakebench} fills the blank in explainable image forensics for forgeries generated by deep generative models (e.g. Diffusion models). ForgerySleuth~\cite{sun2024forgerysleuth} obtains hierarchical forgery description annotation with the proposed Chain-of-Clues for the MIML~\cite{Qu_2024_CVPR}, a high quality image forensics dataset with more than 123k manually forged images and their corresponding pixel-level annotations. It also improves LISA for natural image forensics with noise domain modeling. SIDA~\cite{huang2024sida} achieves explainable forensics for natural image and AIGC image especially on social media scenario, an effective new model is introduced by improving LISA. Despite the progress made, none of the existing work achieves interpretable forensics on another important task, tampered text detection. Due to the tiny size and visual consistency of tampered text~\cite{CVPR2023DocTamper}, natural image forgery detection methods mostly do not work very well for tampered text detection~\cite{luo2024RTM}, leaving massive risks. It is crucial to develop explainable tampered text detection techniques for reliable text image forensics.

\section{ETTD Dataset}

To fill in the data gap for explainable tampered text detection dataset, we construct a large-scale comprehensive dataset called Explainable Tampered Text Detection (ETTD).

\subsection{Text Tampering}
To ensure the comprehensiveness of the proposed data, we collect multilingual document and card images from the Internet and scene text images from the existing datasets (e.g. ICDAR2013~\cite{karatzas2013icdar} and LSVT~\cite{lsvt}. We then forge some of the collected images with the widely-used methods, copy-move and splicing. Poisson Blending~\cite{perez2023poisson} is employed to reduce visual inconsistency. To further improve the data diversity, we manually edit the text with DiffUTE~\cite{chen2024diffute}, a latest diffusion model for realistic tampered text generation. 

\subsection{Anomaly Description Generation}
As shown in Figure~\ref{fig: main-fig}, we leverage the GPT4o to generate the description of both visual and linguistic anomalies caused by text tampering. Given the different features between tampered text and tampered natural objects~\cite{wang2022tic13}, the textual queries in existing works~\cite{xu2024fakeshield} can not work well for tampered text (e.g. "unnatural depth" is usually observed in tampered natural objects but not in tampered text). To address this issue, we propose an elaborate query that inspires the GPT4o to analyze anomalies for tampered text on six major perspectives, covering texture, integration, alignment, edge artifacts, text coherence, font, as shown in Figure~\ref{fig: main-fig}. The detailed query is presented in the Appendix. 

\begin{figure}[t!]
    \centering
    \includegraphics[width=1.0\linewidth]{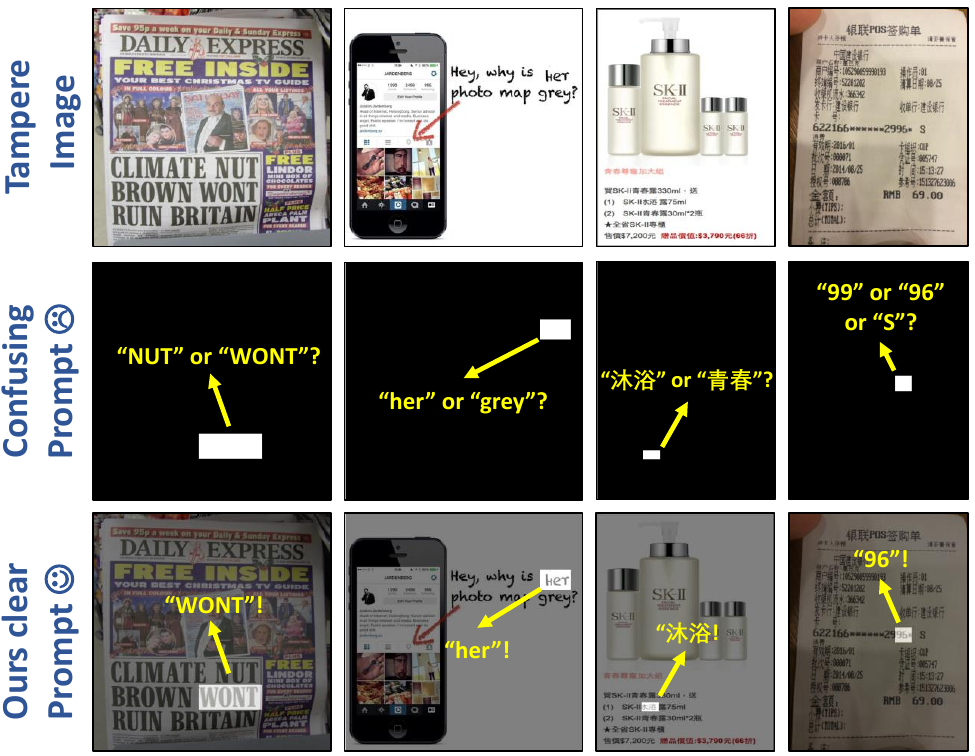}
    \caption{The binary mask prompt as in existing work is confusing in text images. In contrast, our proposed fused mask prompt clearly indicates the content and the exact location of the tampered text.}
    \label{fig: fig3}
    \vspace{-0.4cm}
\end{figure}

We further input this elaborate query along with the tampered image and its corresponding mask annotation into the GPT4o. However, due to the similarity in location and shape of the text instances in an image, directly inputting the binary mask as in existing work~\cite{xu2024fakeshield} will cause considerable confusion to the GPT4o. As shown in Fig.~\ref{fig: fig3}, the annotator model usually struggles to identify the target text with the binary mask, often mistaking a nearby authentic text as a fake text. Analyzing anomaly on authentic text undoubtedly produces incorrect anomaly descriptions. To address this issue, we propose the fused mask prompt, where the original image is fused with the binary mask by pixel-wise weighting. Specifically, given the input image $I\in \mathbb{R}^{H,W,3}$ and the binary mask annotation $M\in \mathbb{B}^{H,W}$, $\mathbb{B} \in \{0,1\}$, the fused mask prompt $M^{fused} \in \mathbb{R}^{H,W,3}$ can be formulated as $M^{fused}=I*\lambda_{1}+M*\lambda_{2}$. We set $\lambda_{1}$ and $\lambda_{2}$ to 0.5 in practical. With the proposed fused mask prompt, the annotator can clearly recognize the tampered text on the target region and better understand where the target region is by referring to the surrounding content. The proposed method significantly reduces hallucination and errors caused by frequent confusion. 

Since the responses of GPT4o are not always correct, directly using the GPT4o responses as annotations leads to poor data quality, while manually verifying the annotation is costly. Our aim is to automatically filter out unsatisfactory responses. We empirically find that, the anomaly description from the GPT4o is also mostly accurate when the GPT4o can correctly recognize the tampered text. This means that the GPT4o is clear about the location of the tampered text and the visual details of it. Inspired by this, we propose to automatically filter out the bad responses with tampered text OCR accuracy~\cite{ICDAR19} lower than 0.8. The OCR ground-truth is obtained from dataset annotation or OCR engine, and is used to replace the GPT4o OCR in the remaining samples to ensure accuracy. The proposed method effectively improves the quality of anomaly description for tampered text in an automatic manner. For authentic text images, the textual description is set to "There is no tampered text in this image.". 

\subsection{Dataset Summary}
As shown in Tab.~\ref{tab: data}, there are 5,500 text images tampered by copy-move, 5,500 text images tampered by splicing and 1,000 text images tampered by DiffUTE in our ETTD dataset. Another 10,500 images without text tampering serve as the authentic part. 20,000 images from the ETTD dataset are split as the training set (ETTD-Train), 1,000 images from the ETTD dataset are split as the test set (ETTD-Test) and another 1,500 images from the ETTD dataset are split as the cross-domain test set (ETTD-CD). The ETTD-CD consists of Copy-move forgeries, Splicing forgeries and authentic images from ICDAR2013, which are not included in ETTD-Train. Therefore, the ETTD-CD has a different data distribution from ETTD-Train and can evaluate model performance on unknown scenarios. Accurate pixel-level annotations for tampered regions are provided to facilitate fine-grained analysis of the tampered text regions. The data hunger of large models can be effectively alleviated with our large-scale diverse data.

\begin{figure*}[t!]
    \centering
    \includegraphics[width=1.0\linewidth]{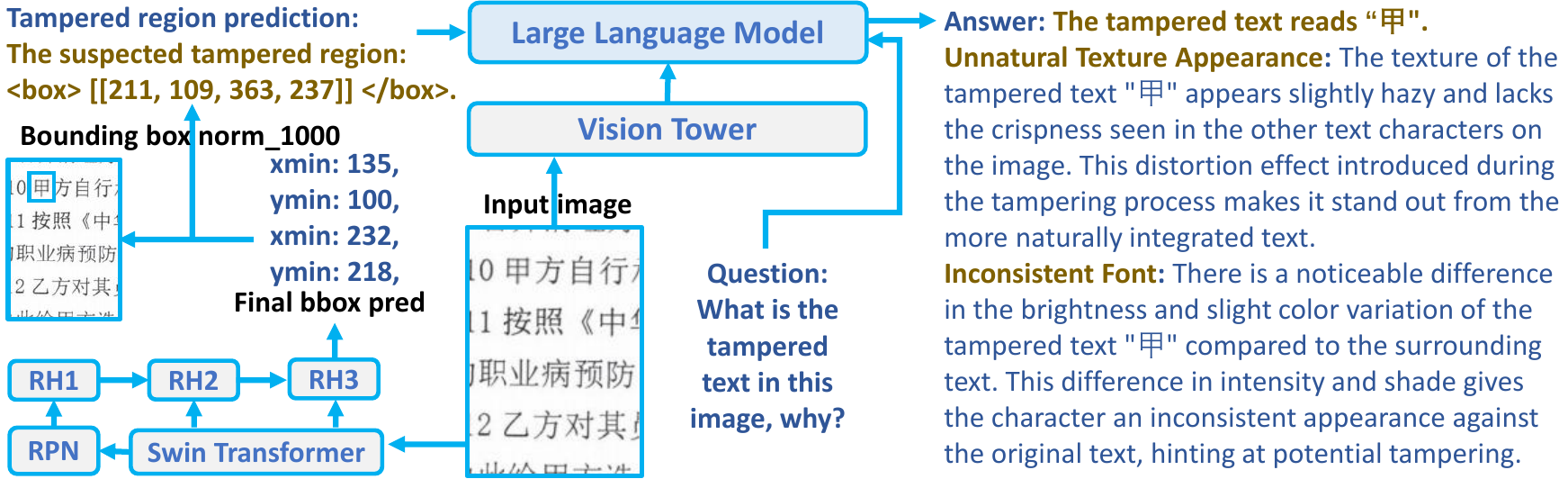}
    \caption{The overall pipeline of the proposed TextSleuth.}
    \label{fig: main-fig2}
    \vspace{-0.2cm}
\end{figure*}

\section{TextSleuth}
The tampered text is mostly tiny in size and the the visual anomaly is mostly unobvious~\cite{luo2024RTM}. Consequently, three major challenges are emerged for interpretable tampered text detection: 1. The multimodal large models suffer from more risks of misidentifying the tampered text, resulting in incorrect anomaly description. 2. A large part of the model parameters are used to find the tampered region, which weakens their ability to analyze and describe the tampered region. 3. The models are more likely to be disturbed by the irrelevant background style, which weakens their generalization on unseen tampering methods and scenarios. To this end, we propose a \textbf{simple-yet-effective} model termed as TextSleuth, which overcomes the above challenges through a two-stage analysis paradigm and a novel reference grounding auxiliary prompt.

As shown in Figure~\ref{fig: main-fig2}, given an input image, the suspected tampered text region is initially detected by a Swin-Transformer based detection model with cascaded RoI heads~\cite{cascade}. The predicted coordinates are then normalized to 0-1000 and are converted to the reference grounding auxiliary prompt "The suspected tampered text $\langle box \rangle [[x_{min}, y_{min}, x_{max}, y_{max}]]\langle /box \rangle $".~Given that the reference grounding comprehension task is involved in the pre-training stage of most large models~\cite{wang2024qwen2vl}, the large language model can effortlessly comprehend the target location in the proposed auxiliary prompt. In the auxiliary prompt, the large language model naturally pays special attention to the region represented by the coordinates, as it has learned in its pre-training stage. This differs from existing work~\cite{li2024forgerygpt} that forces the model to look at the suspected region with binary mask embeddings, which is confusing in indicating tampered text, violates the pre-training paradigm and causes more forgetting. The auxiliary prompt is fed into the large language model along with the image tokens and the original question, to obtain the recognition and describe the anomaly for tampered text.

Despite its simplicity, the proposed method effectively addresses the three major challenges in explainable tampered text detection: 1. The initial prediction of the suspect region significantly reduces the risk of misidentifying the tampered region and reduces hallucination. 2. The detection prior alleviates the difficulty in detecting tampered text, models can save more parameters to obtain better anomaly analysis and description. 3. By focusing on the tampered region, the model gets rid of the interference from unrelated background styles, learns more general features during training, and thereby obtains improved generalisation to unseen tampering methods and scenarios.

\section{Experiments}
We conduct experiments on both the proposed ETTD dataset and the public Tampered IC-13 dataset~\cite{wang2022tic13} with multiple advanced large multimodal models, including GPT4o~\cite{openai2024gpt4technicalreport}, Yi-VL-6B~\cite{yi}, DeepSeekVL-7B~\cite{lu2024deepseekvl}, MiniCPMV2.5~\cite{hu2024minicpm}, the 1B to 8B versions of Intern2VL~\cite{internvl2} and the 2B, 7B versions of Qwen2VL~\cite{wang2024qwen2vl}. We fine-tune all models except GPT4o on the ETTD training set for 5 epochs with the same settings, which is sufficient for all models to achieve their best performance.
\subsection{Evaluation Metric}
To evaluate the similarity between the predicted anomaly description and the textual annotation, we calculate the OCR accuracy~\cite{ICDAR19} for tampered text recognition and the paragraph cosine similarity for non-OCR parts. The weighted summary of OCR accuracy and paragraph similarity is used as the final similarity score. For misclassified samples, the paragraph cosine similarity is set directly to 0 as the gist is opposite.
Specifically, we extract the content within the quotation marks from the first predicted sentence and use it to calculate the OCR accuracy $Acc_{OCR}$. We then remove stop-words and the content within the quotation marks from both prediction $P_{pred}$ and ground-truth paragraphs $P_{gt}$ for more accurate paragraph similarity calculation. The paragraph feature vectors $V_{pred}, V_{gt}$ are obtained by averaging the word vectors in each paragraph, $V_{pred}=average([W2V(word) \enspace for\enspace word \enspace in \enspace P_{pred}])$, $V_{gt}=average([W2V(word) \enspace for\enspace word \enspace in \enspace P_{gt}])$, where $W2V$ is the pretrained word-to-vector function~\cite{w2v}. Finally, the cosine similarity between the two paragraph feature vectors is used as the paragraph similarity score, $Sim_{para}=Cos(V_{pred}, V_{gt})$. We have manually verified that better predictions almost always lead to higher cosine similarity scores.
 The final similarity score $Sim$ is calculated by $Sim=0.3*Acc_{OCR}+0.7*Sim_{para}$. The common accuracy metric~\cite{guillaro2023trufor} is adopted for image forgery classification task. 
 
\subsection{Implement Details}
The vision tower and projector of the large multimodal model are full-parameter fine-tuned and the large language model part is LoRA~\cite{hu2021lora} fine-tuned with rank 8 and alpha 16. We adopt AdamW~\cite{adamw} optimizer with a learning rate decaying linearly from 1e-4 to 0. The batch-size is set to 16 for all models and the experiments are run on NVIDIA A100 80GB GPUs. We set the maximum input area to 1.2M pixels for the Qwen2VL models. In the proposed TextSleuth model, the Swin-Transformer based detection model is trained for 30 epochs on the ETTD training set, with a batch-size of 16 and a maximum input resolution of 1.2M pixels. The AdamW optimizer is employed with a learning rate that decays linearly from 6e-6 to 3e-6.

For all the fine-tuned models, the input text query is "What is the tampered text in this image, why?", which matches the training data. The image classification prediction is regarded as "tampered" if the edit distance between the model output and the string "There is no tampered text in this image." is greater than 3. For GPT4o and other pre-trained models, to output the most similar format with the annotation, the query is set to "Does this image have tampered text on it? Please start your answer with "Yes" or "No". If "Yes", then recognize the tampered text and describe the anomaly of the tampered region.". The image classification prediction is regarded as "tampered" if the output starts with "Yes".

\subsection{Comparison Study}
\noindent \textbf{Anomaly Description}. The comparison results of anomaly description on the ETTD dataset are shown in Table~\ref{tab: main_comp}. Four conclusions can be drawn through analyzing the results:

\noindent (1) \textbf{High input resolution is essential for models to correctly recognize the tampered text and explain its anomaly, but it can also slightly weaken model generalization to unfamiliar scenarios.} On the ETTD-Test set, the Yi-VL-6B with the lowest input resolution 448$\times$448 achieves the lowest final score of 68.5, which is notably lower than the fine-tuned models. Resizing the input image to such a small resolution makes the subtle visual anomaly hard to detect, thus Yi-VL-6B performs the worst. However, the Yi-VL-6B performs much better on ETTD-CD, where the tampered text is mostly larger. Most of the other models suffer significant performance degradation due to the unfamiliar scenario. This indicates that high input resolution mostly weaken model generalization on unfamiliar scenarios.

\noindent (2) \textbf{The scaling law also applies to the explainable tampered text detection task.} Even within the same series (e.g. InternVL2 or Qwen2VL) where the vision tower is the same and the pre-training data is similar, models with larger LLMs mostly perform better. For example, Qwen2VL-7B achieves an average final score of 90.7, which is better than Qwen2VL-2B. This confirms that the scaling law behind our task. 

\noindent (3) \textbf{Model performance on the explainable tampered text detection task is highly related to its pre-trained model performance on general task.} For example, on the OpenCompass, the pre-trained MiniCPMV2.5 model performs better than DeepSeekVL-7B, which is consistent with the model performance of their fine-tuned versions on the explainable forensic task. A better large multimodal model on general task is likely to be more intelligent and can better learn to describe the anomaly of tampered text.

\noindent (4) \textbf{The proposed TextSleuth performs the best on both ETTD-Test and ETTD-CD, significantly outperforming other models in both in-domain and cross-domain scenarios.} This verifies that the proposed auxiliary prompt improves model's fine-grained perception and helps the model to produce high quality anomaly description by focusing its attention on the suspected region at start.

\noindent \textbf{Image Forgery Classification}. The comparison results of the image forgery classification are shown in Tab.~\ref{tab: cls_comp}. The public tampered IC-13 dataset used in evaluation consists of texts tampered by SR-Net~\cite{ACMMM_SRNet} and can also evaluate model's cross domain performance on unknown tampering method. The proposed TextSleuth considerably outperforms existing methods on all the three benchmarks, and improves the Qwen2VL-7B baseline by +3.5 points, +7.9 points and + 12.7 points on the three benchmarks respectively, demonstrating the effectiveness of the proposed method. Similar conclusions can be drawn as analyzed in the above paragraphs.

\noindent \textbf{Auto-annotation}. To verify the effectiveness of the proposed fused mask prompt, we manually obtain the OCR results of 100 random tampered texts from the collected data. We compare the tampered text OCR performance between the binary mask prompt as in existing work and the proposed fused mask prompt, the results are shown in Table~\ref{tab: abl_data}. The annotator GPT4o has significantly higher OCR accuracy with the proposed prompt. The higher OCR accuracy indicates that the large model can better understand the actual location of the tampered text. Therefore, the anomaly description from GPT4o is also more accurate with the method.

\subsection{Ablation Study}
The ablation study of the proposed TextSleuth is shown in Table~\ref{tab: main_abl}. We conduct experiments on three base multimodal models. For each base model (e.g. InternVL2-2B), there are four ablation settings. Setting (1) is the official pre-trained model performance. Setting (2) is the official model fine-tuned on the ETTD Train. Setting (3) is the TextSleuth fine-tuned with the proposed grounding auxiliary prompt. Setting (4) is the TextSleuth with the perfect tampered text detector. Four conclusions can be drawn through analysis: 

\noindent (1) \textbf{The existing multimodal models do not have the ability to recognize tampered text and the anomaly.} All three base models perform poorly in setting (1), but much better in setting (2). This confirms that the official open-source models are mostly incapable of detecting tampered text. Training them on the ETTD data is essential for them to gain the ability for explainbale tampered text detection.

\noindent (2) \textbf{The proposed auxiliary prompt can significantly improve model performance across different base models.} For each of the three base models, the model performance in setting (3) is significantly better than that in setting (2) (+10.8 points average final score for InternVL2-2B, +9.2 for Qwen2VL-2B and +6.5 for Qwen2VL-7B). These improvements are achieved by the proposed two-stage analysis paradigm and the auxiliary prompt in our TextSleuth. The proposed methods alleviate the difficulty in detecting tampered region and make the models better focused on analyzing the anomaly, resulting in an improved anomaly description quality. Additionally, by focusing on the tampered text with the proposed prompt, the models can learn more general features by reducing the interference from unrelated background styles. Consequently, the model's cross-domain generalization is considerably improved. The huge improvements on different basic large multimodal models also demonstrate that our TextSleuth is widely applicable. 

\noindent (3) \textbf{The performance of our TextSleuth can be further improved with better tampered region detectors.} For all of the three base models, model performance in setting (4) is better than those in setting (3). The improvement is achieved by eliminating the errors of the initial tampered text box prediction. Therefore, our TextSleuth can easily be improved in the future with an advanced tampered text region detector.

\smallskip

\noindent \textbf{Robustness Evaluation}. We evaluate the robustness of the TextSleuth under different JPEG compression quality factors and different resize factors on ETTD-Test and ETTD-CD. As shown in Table~\ref{tab: robust}, the stable performance under common distortions has verified the robustness of our TextSleuth.

\smallskip
The \textbf{prediction visualization} is presented in the Appendix.

\begin{table*}[ht!]
\centering
\caption{Comparsion study of the proposed method.}
\vspace{-0.25cm}
\begin{tabular}{ccccccccccccccc}
\hline
\multirow{2}{*}{Methods} &  & \multicolumn{5}{c}{ETTD-Test (in-domain)} &  & \multicolumn{5}{c}{ETTD-CD (cross-domain)} &  & Average \\ \cline{3-7} \cline{9-13} \cline{15-15} 
 &  & \begin{tabular}[c]{@{}c@{}}OCR\\ accuracy\end{tabular} &  & \begin{tabular}[c]{@{}c@{}}Cosine\\ similarity\end{tabular} &  & \begin{tabular}[c]{@{}c@{}}Final\\ score\end{tabular} &  & \begin{tabular}[c]{@{}c@{}}OCR\\ accuracy\end{tabular} &  & \begin{tabular}[c]{@{}c@{}}Cosine\\ similarity\end{tabular} &  & \begin{tabular}[c]{@{}c@{}}Final\\ score\end{tabular} &  & \begin{tabular}[c]{@{}c@{}}Final\\ score\end{tabular} \\ \cline{1-1} \cline{3-3} \cline{5-5} \cline{7-7} \cline{9-9} \cline{11-11} \cline{13-13} \cline{15-15} 
GPT4o &  & 48.3 &  & 66.1 &  & 60.7 &  & 74.6 &  & 78.0 &  & 77.0 &  & 68.9 \\
Yi-VL-6B &  & 49.9 &  & 76.5 &  & 68.5 &  & 64.3 &  & 81.4 &  & 76.2 &  & 72.4 \\
DeepSeekVL-7B &  & 66.6 &  & 86.9 &  & 80.8 &  & 37.9 &  & 64.7 &  & 56.7 &  & 68.8 \\
MiniCPMV2.5 &  & 79.3 &  & 92.6 &  & 88.6 &  & 68.9 &  & 74.8 &  & 73.0 &  & 80.8 \\
InternVL2-1B &  & 77.8 &  & 89.1 &  & 85.7 &  & 79.2 &  & 84.0 &  & 82.5 &  & 84.1 \\
InternVL2-2B &  & 81.1 &  & 91.5 &  & 88.3 &  & 78.2 &  & 82.7 &  & 81.3 &  & 84.8 \\
InternVL2-4B &  & 75.8 &  & 82.4 &  & 80.4 &  & 91.4 &  & 94.0 &  & 93.1 &  & 86.8 \\
InternVL2-8B &  & 80.9 &  & 90.7 &  & 87.7 &  & 80.0 &  & 85.1 &  & 83.5 &  & 85.6 \\
Qwen2VL-2B &  & 84.8 &  & 93.7 &  & 91.0 &  & 82.1 &  & 85.0 &  & 84.1 &  & 87.6 \\
Qwen2VL-7B &  & 87.1 &  & 94.8 &  & 92.4 &  & 87.1 &  & 89.9 &  & 88.9 &  & 90.7 \\
TextSleuth-7B (Ours) &  & \textbf{92.6} & \textbf{} & \textbf{98.3} & \textbf{} & \textbf{96.5} & \textbf{} & \textbf{97.7} & \textbf{} & \textbf{98.1} & \textbf{} & \textbf{97.9} & \textbf{} & \textbf{97.2} \\ \hline
\end{tabular}
\label{tab: main_comp}
\vspace{-0.15cm}
\end{table*}

\begin{table*}[ht!]
\centering
\setlength{\tabcolsep}{2pt}
\caption{Ablation study of the proposed method. "SFT" denotes surprised fine-tuning. "TextSleuth" denotes equipping the model with the proposed TextSleuth method. "Perfect Detector" denotes using ground-truth tampered region boxes in the TextSleuth's auxiliary prompt.}
\vspace{-0.25cm}
\begin{tabular}{ccccccccccccccccccccccc}
\hline
\multirow{2}{*}{\begin{tabular}[c]{@{}c@{}}Base\\ Multi-modal\\ Model\end{tabular}} &  & \multicolumn{7}{c}{Ablation settings} &  & \multicolumn{5}{c}{ETTD-Test (in-domain)} &  & \multicolumn{5}{c}{ETTD-CD (cross-domain)} &  & Average \\ \cline{3-9} \cline{11-15} \cline{17-21} \cline{23-23} 
 &  & Num &  & SFT &  & \begin{tabular}[c]{@{}c@{}}Text\\ Sleuth\end{tabular} &  & \begin{tabular}[c]{@{}c@{}}Perfect\\ Detector\end{tabular} &  & \begin{tabular}[c]{@{}c@{}}OCR\\ accuracy\end{tabular} &  & \begin{tabular}[c]{@{}c@{}}Cosine\\ similarity\end{tabular} &  & \begin{tabular}[c]{@{}c@{}}Final\\ score\end{tabular} &  & \begin{tabular}[c]{@{}c@{}}OCR\\ accuracy\end{tabular} &  & \begin{tabular}[c]{@{}c@{}}Cosine\\ similarity\end{tabular} &  & \begin{tabular}[c]{@{}c@{}}Final\\ score\end{tabular} &  & \begin{tabular}[c]{@{}c@{}}Final\\ score\end{tabular} \\ \cline{1-1} \cline{3-3} \cline{5-5} \cline{7-7} \cline{9-9} \cline{11-11} \cline{13-13} \cline{15-15} \cline{17-17} \cline{19-19} \cline{21-21} \cline{23-23} 
\multirow{4}{*}{InternVL2-2B} &  & (1) &  & $\times$ &  & $\times$ &  & $\times$ &  & 14.1 &  & 57.8 &  & 44.7 &  & 34.3 &  & 64.4 &  & 55.4 &  & 50.1 \\
 &  & (2) &  & \checkmark &  & $\times$ &  & $\times$ &  & 81.1 &  & 91.5 &  & 88.3 &  & 78.2 &  & 82.7 &  & 81.3 &  & 84.8 \\
 &  & (3) &  & \checkmark &  & \checkmark &  & $\times$ &  & 83.1 &  & 98.3 &  & 93.6 &  & 96.5 &  & 98.1 &  & 97.5 &  & 95.6 \\
 &  & (4) &  & \checkmark &  & \checkmark &  & \checkmark &  & 83.7 &  & 99.4 &  & 94.6 &  & 97.7 &  & 99.4 &  & 98.8 &  & 96.7 \\ \cline{1-1} \cline{3-9} \cline{11-15} \cline{17-21} \cline{23-23} 
\multirow{4}{*}{Qwen2VL-2B} &  & (1) &  & $\times$ &  & $\times$ &  & $\times$ &  & 18.5 &  & 57.0 &  & 45.5 &  & 29.8 &  & 63.8 &  & 53.6 &  & 49.6 \\
 &  & (2) &  & \checkmark &  & $\times$ &  & $\times$ &  & 84.8 &  & 93.7 &  & 91.0 &  & 82.1 &  & 85.0 &  & 84.1 &  & 87.6 \\
 &  & (3) &  & \checkmark &  & \checkmark &  & $\times$ &  & 90.4 &  & 98.2 &  & 95.8 &  & 97.2 &  & 98.0 &  & 97.7 &  & 96.8 \\
 &  & (4) &  & \checkmark &  & \checkmark &  & \checkmark &  & 91.3 &  & 99.3 &  & 96.8 &  & 98.5 &  & 99.3 &  & 99.0 &  & 97.9 \\ \cline{1-1} \cline{3-9} \cline{11-15} \cline{17-21} \cline{23-23} 
\multirow{4}{*}{Qwen2VL-7B} &  & (1) &  & $\times$ &  & $\times$ &  & $\times$ &  & 14.0 &  & 41.8 &  & 33.5 &  & 36.4 &  & 53.4 &  & 48.3 &  & 40.9 \\
 &  & (2) &  & \checkmark &  & $\times$ &  & $\times$ &  & 87.1 &  & 94.8 &  & 92.4 &  & 87.1 &  & 89.9 &  & 88.9 &  & 90.7 \\
 &  & (3) &  & \checkmark &  & \checkmark &  & $\times$ &  & 92.6 &  & 98.3 &  & 96.5 &  & 97.7 &  & 98.1 &  & 97.9 &  & 97.2 \\
 &  & (4) &  & \checkmark &  & \checkmark &  & \checkmark &  & 93.6 &  & 99.4 &  & 97.6 &  & 99.0 &  & 99.4 &  & 99.2 &  & 98.4 \\ \hline
\end{tabular}
\label{tab: main_abl}
\vspace{-0.35cm}
\end{table*}

\begin{table}[ht!]
\centering
\vspace{+0.2cm}
\setlength{\tabcolsep}{2pt}
\caption{Accuracy performance of different large multimodal models on image forgery classification task.}
\vspace{-0.25cm}
\begin{tabular}{ccccccc}
\hline
\multirow{3}{*}{Method} &  & \multirow{3}{*}{\begin{tabular}[c]{@{}c@{}}ETTD-\\ Test\\ (in-domain)\end{tabular}} &  & \multirow{3}{*}{\begin{tabular}[c]{@{}c@{}}ETTD-\\ CD\\ (out-domain)\end{tabular}} &  & \multirow{3}{*}{\begin{tabular}[c]{@{}c@{}}Tampered-\\ IC13\\ (zero-shot)\end{tabular}} \\
 & \multicolumn{1}{l}{} &  & \multicolumn{1}{l}{} &  & \multicolumn{1}{l}{} &  \\
 &  &  &  &  & \multicolumn{1}{l}{} &  \\ \cline{1-1} \cline{3-3} \cline{5-5} \cline{7-7} 
GPT4o &  & 67.3 &  & 79.3 & \textbf{} & 82.8 \\
Yi-VL-6B &  & 76.9 & \textbf{} & 81.9 & \textbf{} & 45.9 \\
DeepSeekVL-7B &  & 87.4 & \textbf{} & 66.7 & \textbf{} & 76.4 \\
MiniCPMV2.5 &  & 93.2 &  & 75.5 & \textbf{} & 56.7 \\
InternVL2-1B &  & 89.7 &  & 84.6 & \textbf{} & 59.2 \\
InternVL2-2B & \multicolumn{1}{l}{} & 92.1 &  & 83.3 & \textbf{} & 58.8 \\
InternVL2-4B & \multicolumn{1}{l}{} & 82.8 &  & 94.5 & \textbf{} & 36.1 \\
InternVL2-8B & \multicolumn{1}{l}{} & 91.2 &  & 85.7 & \textbf{} & 60.5 \\
Qwen2VL-2B &  & 94.3 &  & 85.7 & \textbf{} & 73.8 \\
Qwen2VL-7B &  & 95.4 & \textbf{} & 90.5 &  & 75.1 \\
TextSleuth-7B &  & \textbf{98.9} & \textbf{} & \textbf{98.6} &  & \textbf{88.4} \\ \hline
\end{tabular}
\label{tab: cls_comp}
\end{table}

\begin{table}[t!]
\centering
\setlength{\tabcolsep}{2pt}
\caption{Comparison study for the proposed fused mask prompt.}
\vspace{-0.25cm}
\begin{tabular}{ccccc}
\hline
\multirow{2}{*}{Method} &  & \multirow{2}{*}{\begin{tabular}[c]{@{}c@{}}OCR\\ Accuracy\end{tabular}} &  & \multirow{2}{*}{\begin{tabular}[c]{@{}c@{}}Perfect\\ Match\end{tabular}} \\
 &  &  &  &  \\ \cline{1-1} \cline{3-3} \cline{5-5} 
Binary mask prompt &  & 47.3 &  & 30.4 \\
Fused mask prompt (Ours) &  & \textbf{84.2} &  & \textbf{73.0} \\ \hline
\end{tabular}
\label{tab: abl_data}
\vspace{+0.2cm}
\end{table}

\section{Conclusion}
This paper is the first work that achieves explainable tampered text detection, by describing the anomaly of the tampered text image with natural language. Given the absence of dataset for this task, we construct a large-scale comprehensive dataset termed as ETTD, which consists of multilingual document and scene text images tampered by copy-move, splicing and AIGC-based text editing. Given the obtained tampered text, we obtain the anomaly description by prompting GPT4o with elaborate queries, the tampered image and tampered region annotation. However, due to the similar shape and position of the text, prompting GPT4o with a binary mask as in existing works mostly confuses the GPT4o and leads to incorrect responses. To address this issue, we propose to prompt GPT4o with a fused mask, which is obtained by weighting the image with the binary mask. Experiments verify that the proposed fused mask prompt significantly improves the annotation quality. Moreover, given that the incorrect recognition of tampered text means unclear perception and leads to bad anomaly description, we also propose to filter out the responses with low tampered text OCR accuracy, which can effectively improve annotation quality in an automatic manner. The proposed ETTD dataset has both in-domain and cross-domain test subsets, which allows a thorough evaluation of model generalization. In addition, a novel TextSleuth model is proposed to further improve explainable tampered text detection, which overcomes several major challenges in the field with a two-stage analysis paradigm and an auxiliary prompt. Experiments have confirmed that the proposed method considerably improves upon different baseline models, and that our TextSleuth notably outperforms existing methods in both in-domain and cross-domain evaluation on both the ETTD and public datasets. In-depth analysis is also provided to inspire further work. We believe that our \textbf{valuable} ETTD dataset and our \textbf{first-of-its-kind, simple-yet-effective} methods can shed light on the further research on interpretable tampered text detection.  

\begin{table}[t!]
\centering
\setlength{\tabcolsep}{1pt}
\vspace{+0.2cm}
\caption{Robustness evaluation under common distortions.}
\vspace{-0.25cm}
\begin{tabular}{ccccccccccc}
\hline
\multirow{3}{*}{\begin{tabular}[c]{@{}c@{}}Average\\ final\\ score\end{tabular}} &  & \multirow{3}{*}{Ori.} &  & \multirow{3}{*}{\begin{tabular}[c]{@{}c@{}}JPEG\\ compress\\ quality75\end{tabular}} &  & \multirow{3}{*}{\begin{tabular}[c]{@{}c@{}}JPEG\\ compress\\ quality50\end{tabular}} &  & \multirow{3}{*}{\begin{tabular}[c]{@{}c@{}}Image\\ resize\\ factor0.75\end{tabular}} &  & \multirow{3}{*}{\begin{tabular}[c]{@{}c@{}}Image\\ resize\\ factor0.5\end{tabular}} \\
 &  &  &  &  &  &  &  &  &  &  \\
 &  &  &  &  &  &  &  &  &  &  \\ \cline{1-1} \cline{3-3} \cline{5-5} \cline{7-7} \cline{9-9} \cline{11-11} 
Qwen2VL &  & 90.7 &  & 89.6 &  & 87.2 &  & 89.2 &  & 86.1 \\
TextSleuth &  & \textbf{97.2} &  & \textbf{96.3} &  & \textbf{94.4} &  & \textbf{95.8} &  & \textbf{93.0} \\
\hline
\end{tabular}
\label{tab: robust}
\vspace{-0.5cm}
\end{table}

\bibliographystyle{named}
\bibliography{ijcai24}

\begin{thebibliography}{}

\bibitem[\protect\citeauthoryear{Ahmed and Shafait}{2014}]{ahmed2014forgery}
Amr Gamal~Hamed Ahmed and Faisal Shafait.
\newblock Forgery detection based on intrinsic document contents.
\newblock In {\em 2014 11th IAPR International Workshop on Document Analysis Systems}, pages 252--256. IEEE, 2014.

\bibitem[\protect\citeauthoryear{AI \bgroup \em et al.\egroup }{2024}]{yi}
01. AI, :, Alex Young, Bei Chen, Chao Li, Chengen Huang, Ge~Zhang, Guanwei Zhang, Heng Li, Jiangcheng Zhu, Jianqun Chen, Jing Chang, Kaidong Yu, Peng Liu, Qiang Liu, Shawn Yue, Senbin Yang, Shiming Yang, Tao Yu, Wen Xie, Wenhao Huang, Xiaohui Hu, Xiaoyi Ren, Xinyao Niu, Pengcheng Nie, Yuchi Xu, Yudong Liu, Yue Wang, Yuxuan Cai, Zhenyu Gu, Zhiyuan Liu, and Zonghong Dai.
\newblock Yi: Open foundation models by 01.ai, 2024.

\bibitem[\protect\citeauthoryear{{Alibaba Security}}{2020}]{sacp}
{Alibaba Security}.
\newblock Security ai challenger program.
\newblock \url{https://tianchi.aliyun.com/competition/entrance/531812/introduction}, 2020.

\bibitem[\protect\citeauthoryear{Cai and Vasconcelos}{2018}]{cascade}
Zhaowei Cai and Nuno Vasconcelos.
\newblock Cascade r-cnn: Delving into high quality object detection.
\newblock In {\em Proceedings of the IEEE Conference on Computer Vision and Pattern Recognition (CVPR)}, June 2018.

\bibitem[\protect\citeauthoryear{Chen \bgroup \em et al.\egroup }{2024a}]{chen2024cma}
Changsheng Chen, Liangwei Lin, Yongqi Chen, Bin Li, Jishen Zeng, and Jiwu Huang.
\newblock Cma: A chromaticity map adapter for robust detection of screen-recapture document images.
\newblock In {\em Proceedings of the IEEE/CVF Conference on Computer Vision and Pattern Recognition}, pages 15577--15586, 2024.

\bibitem[\protect\citeauthoryear{Chen \bgroup \em et al.\egroup }{2024b}]{chen2024diffute}
Haoxing Chen, Zhuoer Xu, Zhangxuan Gu, Yaohui Li, Changhua Meng, Huijia Zhu, Weiqiang Wang, et~al.
\newblock Diffute: Universal text editing diffusion model.
\newblock {\em Advances in Neural Information Processing Systems}, 36, 2024.

\bibitem[\protect\citeauthoryear{Chen \bgroup \em et al.\egroup }{2024c}]{internvl2}
Zhe Chen, Weiyun Wang, Hao Tian, Shenglong Ye, Zhangwei Gao, Erfei Cui, Wenwen Tong, Kongzhi Hu, Jiapeng Luo, Zheng Ma, et~al.
\newblock How far are we to gpt-4v? closing the gap to commercial multimodal models with open-source suites.
\newblock {\em arXiv preprint arXiv:2404.16821}, 2024.

\bibitem[\protect\citeauthoryear{Dong \bgroup \em et al.\egroup }{2024}]{dong2024robust}
Li~Dong, Weipeng Liang, and Rangding Wang.
\newblock Robust text image tampering localization via forgery traces enhancement and multiscale attention.
\newblock {\em IEEE Transactions on Consumer Electronics}, 2024.

\bibitem[\protect\citeauthoryear{Guillaro \bgroup \em et al.\egroup }{2023}]{guillaro2023trufor}
Fabrizio Guillaro, Davide Cozzolino, Avneesh Sud, Nicholas Dufour, and Luisa Verdoliva.
\newblock Trufor: Leveraging all-round clues for trustworthy image forgery detection and localization.
\newblock In {\em Proceedings of the IEEE/CVF Conference on Computer Vision and Pattern Recognition}, pages 20606--20615, 2023.

\bibitem[\protect\citeauthoryear{Hu \bgroup \em et al.\egroup }{2021}]{hu2021lora}
Edward~J Hu, Yelong Shen, Phillip Wallis, Zeyuan Allen-Zhu, Yuanzhi Li, Shean Wang, Lu~Wang, and Weizhu Chen.
\newblock Lora: Low-rank adaptation of large language models.
\newblock {\em arXiv preprint arXiv:2106.09685}, 2021.

\bibitem[\protect\citeauthoryear{Hu \bgroup \em et al.\egroup }{2024}]{hu2024minicpm}
Shengding Hu, Yuge Tu, Xu~Han, Chaoqun He, Ganqu Cui, Xiang Long, Zhi Zheng, Yewei Fang, Yuxiang Huang, Weilin Zhao, Xinrong Zhang, Zheng~Leng Thai, Kaihuo Zhang, Chongyi Wang, Yuan Yao, Chenyang Zhao, Jie Zhou, Jie Cai, Zhongwu Zhai, Ning Ding, Chao Jia, Guoyang Zeng, Dahai Li, Zhiyuan Liu, and Maosong Sun.
\newblock Minicpm: Unveiling the potential of small language models with scalable training strategies, 2024.

\bibitem[\protect\citeauthoryear{Huang \bgroup \em et al.\egroup }{2024a}]{huang2024ffaa}
Zhengchao Huang, Bin Xia, Zicheng Lin, Zhun Mou, Wenming Yang, and Jiaya Jia.
\newblock Ffaa: Multimodal large language model based explainable open-world face forgery analysis assistant, 2024.

\bibitem[\protect\citeauthoryear{Huang \bgroup \em et al.\egroup }{2024b}]{huang2024sida}
Zhenglin Huang, Jinwei Hu, Xiangtai Li, Yiwei He, Xingyu Zhao, Bei Peng, Baoyuan Wu, Xiaowei Huang, and Guangliang Cheng.
\newblock Sida: Social media image deepfake detection, localization and explanation with large multimodal model, 2024.

\bibitem[\protect\citeauthoryear{Karatzas \bgroup \em et al.\egroup }{2013}]{karatzas2013icdar}
Dimosthenis Karatzas, Faisal Shafait, Seiichi Uchida, Masakazu Iwamura, Lluis~Gomez i~Bigorda, Sergi~Robles Mestre, Joan Mas, David~Fernandez Mota, Jon~Almazan Almazan, and Lluis~Pere De~Las~Heras.
\newblock Icdar 2013 robust reading competition.
\newblock In {\em 2013 12th international conference on document analysis and recognition}, pages 1484--1493. IEEE, 2013.

\bibitem[\protect\citeauthoryear{Lai \bgroup \em et al.\egroup }{2024}]{lai2024lisa}
Xin Lai, Zhuotao Tian, Yukang Chen, Yanwei Li, Yuhui Yuan, Shu Liu, and Jiaya Jia.
\newblock Lisa: Reasoning segmentation via large language model.
\newblock In {\em Proceedings of the IEEE/CVF Conference on Computer Vision and Pattern Recognition}, pages 9579--9589, 2024.

\bibitem[\protect\citeauthoryear{Lampert \bgroup \em et al.\egroup }{2006}]{lampert2006printing}
Christoph~H Lampert, Lin Mei, and Thomas~M Breuel.
\newblock Printing technique classification for document counterfeit detection.
\newblock In {\em 2006 International Conference on Computational Intelligence and Security}, volume~1, pages 639--644. IEEE, 2006.

\bibitem[\protect\citeauthoryear{Li \bgroup \em et al.\egroup }{2024a}]{li2024forgerygpt}
Jiawei Li, Fanrui Zhang, Jiaying Zhu, Esther Sun, Qiang Zhang, and Zheng-Jun Zha.
\newblock Forgerygpt: Multimodal large language model for explainable image forgery detection and localization, 2024.

\bibitem[\protect\citeauthoryear{Li \bgroup \em et al.\egroup }{2024b}]{li2024fakebench}
Yixuan Li, Xuelin Liu, Xiaoyang Wang, Bu~Sung Lee, Shiqi Wang, Anderson Rocha, and Weisi Lin.
\newblock Fakebench: Probing explainable fake image detection via large multimodal models, 2024.

\bibitem[\protect\citeauthoryear{Lian \bgroup \em et al.\egroup }{2024}]{mmtt}
Jingchun Lian, Lingyu Liu, Yaxiong Wang, Yujiao Wu, and Zhedong Zheng.
\newblock A large-scale interpretable multi-modality benchmark for image forgery localization, 2024.

\bibitem[\protect\citeauthoryear{Loshchilov and Hutter}{2017}]{adamw}
Ilya Loshchilov and Frank Hutter.
\newblock Decoupled weight decay regularization.
\newblock {\em arXiv preprint arXiv:1711.05101}, 2017.

\bibitem[\protect\citeauthoryear{Lu \bgroup \em et al.\egroup }{2024}]{lu2024deepseekvl}
Haoyu Lu, Wen Liu, Bo~Zhang, Bingxuan Wang, Kai Dong, Bo~Liu, Jingxiang Sun, Tongzheng Ren, Zhuoshu Li, Hao Yang, Yaofeng Sun, Chengqi Deng, Hanwei Xu, Zhenda Xie, and Chong Ruan.
\newblock Deepseek-vl: Towards real-world vision-language understanding, 2024.

\bibitem[\protect\citeauthoryear{Luo \bgroup \em et al.\egroup }{2024}]{luo2024RTM}
Dongliang Luo, Yuliang Liu, Rui Yang, Xianjin Liu, Jishen Zeng, Yu~Zhou, and Xiang Bai.
\newblock Toward real text manipulation detection: New dataset and new solution.
\newblock {\em Pattern Recognition}, page 110828, 2024.

\bibitem[\protect\citeauthoryear{Mikolov \bgroup \em et al.\egroup }{2018}]{w2v}
Tomas Mikolov, Edouard Grave, Piotr Bojanowski, Christian Puhrsch, and Armand Joulin.
\newblock Advances in pre-training distributed word representations.
\newblock In {\em Proceedings of the International Conference on Language Resources and Evaluation (LREC 2018)}, 2018.

\bibitem[\protect\citeauthoryear{Nayef \bgroup \em et al.\egroup }{2017}]{nayef2017icdar2017}
Nibal Nayef, Fei Yin, Imen Bizid, Hyunsoo Choi, Yuan Feng, Dimosthenis Karatzas, Zhenbo Luo, Umapada Pal, Christophe Rigaud, Joseph Chazalon, et~al.
\newblock Icdar2017 robust reading challenge on multi-lingual scene text detection and script identification-rrc-mlt.
\newblock In {\em 2017 14th IAPR international conference on document analysis and recognition (ICDAR)}, volume~1, pages 1454--1459. IEEE, 2017.

\bibitem[\protect\citeauthoryear{Nguyen \bgroup \em et al.\egroup }{2024}]{nguyen2024editscout}
Quang Nguyen, Truong Vu, Trong-Tung Nguyen, Yuxin Wen, Preston~K Robinette, Taylor~T Johnson, Tom Goldstein, Anh Tran, and Khoi Nguyen.
\newblock Editscout: Locating forged regions from diffusion-based edited images with multimodal llm, 2024.

\bibitem[\protect\citeauthoryear{OpenAI}{2024}]{openai2024gpt4technicalreport}
OpenAI.
\newblock Gpt-4 technical report, 2024.

\bibitem[\protect\citeauthoryear{P{\'e}rez \bgroup \em et al.\egroup }{2023}]{perez2023poisson}
Patrick P{\'e}rez, Michel Gangnet, and Andrew Blake.
\newblock Poisson image editing.
\newblock In {\em Seminal Graphics Papers: Pushing the Boundaries, Volume 2}, pages 577--582. 2023.

\bibitem[\protect\citeauthoryear{Qu \bgroup \em et al.\egroup }{2023}]{CVPR2023DocTamper}
Chenfan Qu, Chongyu Liu, Yuliang Liu, Xinhong Chen, Dezhi Peng, Fengjun Guo, and Lianwen Jin.
\newblock Towards robust tampered text detection in document image: new dataset and new solution.
\newblock In {\em Proceedings of the IEEE/CVF Conference on Computer Vision and Pattern Recognition}, pages 5937--5946, 2023.

\bibitem[\protect\citeauthoryear{Qu \bgroup \em et al.\egroup }{2024a}]{qu2024generalizedtamperedscenetext}
Chenfan Qu, Yiwu Zhong, Fengjun Guo, and Lianwen Jin.
\newblock Generalized tampered scene text detection in the era of generative ai, 2024.

\bibitem[\protect\citeauthoryear{Qu \bgroup \em et al.\egroup }{2024b}]{qu2024omniimlunifiedimagemanipulation}
Chenfan Qu, Yiwu Zhong, Fengjun Guo, and Lianwen Jin.
\newblock Omni-iml: Towards unified image manipulation localization, 2024.

\bibitem[\protect\citeauthoryear{Qu \bgroup \em et al.\egroup }{2024c}]{Qu_2024_CVPR}
Chenfan Qu, Yiwu Zhong, Chongyu Liu, Guitao Xu, Dezhi Peng, Fengjun Guo, and Lianwen Jin.
\newblock Towards modern image manipulation localization: A large-scale dataset and novel methods.
\newblock In {\em Proceedings of the IEEE/CVF Conference on Computer Vision and Pattern Recognition (CVPR)}, pages 10781--10790, June 2024.

\bibitem[\protect\citeauthoryear{Ren \bgroup \em et al.\egroup }{2015}]{ren2015fasterrcnn}
Shaoqing Ren, Kaiming He, Ross Girshick, and Jian Sun.
\newblock Faster r-cnn: Towards real-time object detection with region proposal networks.
\newblock {\em Advances in neural information processing systems}, 28, 2015.

\bibitem[\protect\citeauthoryear{Shao \bgroup \em et al.\egroup }{2023}]{shao2023progressive}
Huiru Shao, Kaizhu Huang, Wei Wang, Xiaowei Huang, and Qiufeng Wang.
\newblock Progressive supervision for tampering localization in document images.
\newblock In {\em International Conference on Neural Information Processing}, pages 140--151. Springer, 2023.

\bibitem[\protect\citeauthoryear{Shao \bgroup \em et al.\egroup }{2025}]{shao2025delving}
Huiru Shao, Zhuang Qian, Kaizhu Huang, Wei Wang, Xiaowei Huang, and Qiufeng Wang.
\newblock Delving into adversarial robustness on document tampering localization.
\newblock In {\em European Conference on Computer Vision}, pages 290--306. Springer, 2025.

\bibitem[\protect\citeauthoryear{Song \bgroup \em et al.\egroup }{2024}]{song2024cross}
Yalin Song, Wenbin Jiang, Xiuli Chai, Zhihua Gan, Mengyuan Zhou, and Lei Chen.
\newblock Cross-attention based two-branch networks for document image forgery localization in the metaverse.
\newblock {\em ACM Transactions on Multimedia Computing, Communications and Applications}, 2024.

\bibitem[\protect\citeauthoryear{Sun \bgroup \em et al.\egroup }{2019}]{lsvt}
Yipeng Sun, Zihan Ni, Chee-Kheng Chng, Yuliang Liu, Canjie Luo, Chun~Chet Ng, Junyu Han, Errui Ding, Jingtuo Liu, Dimosthenis Karatzas, et~al.
\newblock Icdar 2019 competition on large-scale street view text with partial labeling-rrc-lsvt.
\newblock In {\em 2019 International Conference on Document Analysis and Recognition (ICDAR)}, pages 1557--1562. IEEE, 2019.

\bibitem[\protect\citeauthoryear{Sun \bgroup \em et al.\egroup }{2024}]{sun2024forgerysleuth}
Zhihao Sun, Haoran Jiang, Haoran Chen, Yixin Cao, Xipeng Qiu, Zuxuan Wu, and Yu-Gang Jiang.
\newblock Forgerysleuth: Empowering multimodal large language models for image manipulation detection, 2024.

\bibitem[\protect\citeauthoryear{Wang \bgroup \em et al.\egroup }{2022}]{wang2022tic13}
Yuxin Wang, Hongtao Xie, Mengting Xing, Jing Wang, Shenggao Zhu, and Yongdong Zhang.
\newblock Detecting tampered scene text in the wild.
\newblock In {\em European Conference on Computer Vision}, pages 215--232. Springer, 2022.

\bibitem[\protect\citeauthoryear{Wang \bgroup \em et al.\egroup }{2024}]{wang2024qwen2vl}
Peng Wang, Shuai Bai, Sinan Tan, Shijie Wang, Zhihao Fan, Jinze Bai, Keqin Chen, Xuejing Liu, Jialin Wang, Wenbin Ge, Yang Fan, Kai Dang, Mengfei Du, Xuancheng Ren, Rui Men, Dayiheng Liu, Chang Zhou, Jingren Zhou, and Junyang Lin.
\newblock Qwen2-vl: Enhancing vision-language model's perception of the world at any resolution, 2024.

\bibitem[\protect\citeauthoryear{Wu \bgroup \em et al.\egroup }{2019}]{ACMMM_SRNet}
Liang Wu, Chengquan Zhang, Jiaming Liu, Junyu Han, Jingtuo Liu, Errui Ding, and Xiang Bai.
\newblock Editing text in the wild.
\newblock In {\em Proceedings of the 27th ACM International Conference on Multimedia}, MM '19, page 1500–1508, New York, NY, USA, 2019. Association for Computing Machinery.

\bibitem[\protect\citeauthoryear{Xu \bgroup \em et al.\egroup }{2024}]{xu2024fakeshield}
Zhipei Xu, Xuanyu Zhang, Runyi Li, Zecheng Tang, Qing Huang, and Jian Zhang.
\newblock Fakeshield: Explainable image forgery detection and localization via multi-modal large language models, 2024.

\bibitem[\protect\citeauthoryear{Zhang \bgroup \em et al.\egroup }{2019}]{ICDAR19}
Rui Zhang, Yongsheng Zhou, Qianyi Jiang, Qi~Song, Nan Li, Kai Zhou, Lei Wang, Dong Wang, Minghui Liao, Mingkun Yang, et~al.
\newblock Icdar 2019 robust reading challenge on reading chinese text on signboard.
\newblock In {\em 2019 international conference on document analysis and recognition (ICDAR)}, pages 1577--1581. IEEE, 2019.

\end{thebibliography}

\newpage

\begin{figure*}[ht!]
 \centering
 \includegraphics[width=1.0\textwidth]{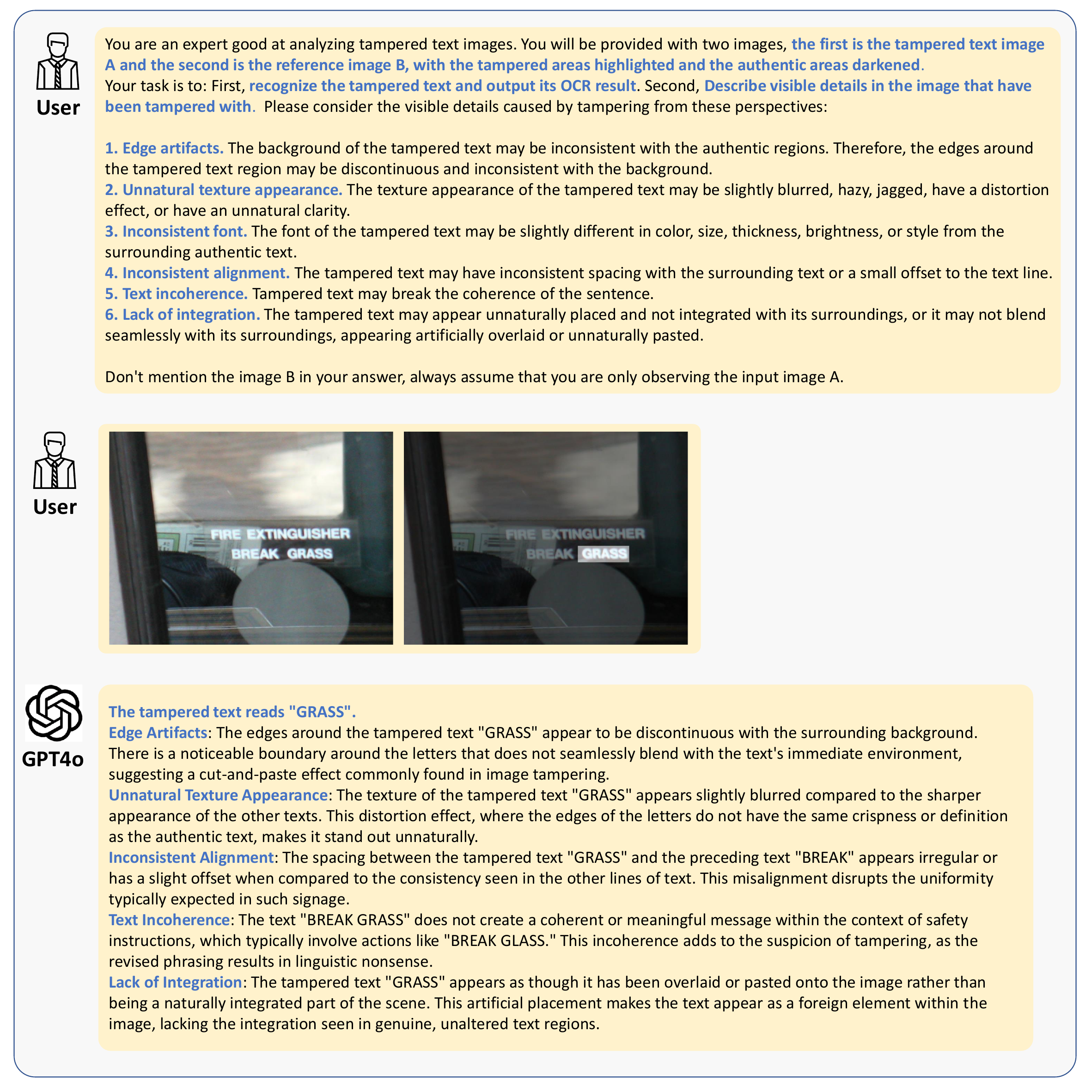}
 \caption{Our proposed textual prompt are specially designed for tampered text can can guide GPT4o to generate high-quality anomaly descriptions for tampered text.}
 \label{fig: prompt}
 \vspace{-0.3cm}
 \end{figure*}

\section{The Proposed Textual Prompt}
Due to the different characteristics of tampered text, existing textual prompts designed for natural objects or deepfakes cannot be directly used to generate high-quality anomaly descriptions for tampered text. To this end, we redesign the textual prompt by summarizing the possible anomalies caused by text tampering into six major perspectives and providing a detailed explanation for each of them.

\smallskip

\noindent The full version of our textual prompt is:

\smallskip

\noindent You are an expert good at analyzing tampered text images. You will be provided with two images, \textbf{the first is the tampered text image A and the second is the reference image B, with the tampered areas highlighted and the authentic areas darkened}. 

\noindent Your task is to: \textbf{First, recognize the tampered text and output its OCR result. Second, Describe visible details in the image that have been tampered with.}  Please consider the visible details caused by tampering from these perspectives.

\noindent \textbf{1. Edge artifacts}. The background of the tampered text may be inconsistent with the authentic regions. Therefore, the edges around the tampered text region may be discontinuous and inconsistent with the background. 

\noindent \textbf{2. Unnatural texture appearance}. The texture appearance of the tampered text may be slightly blurred, hazy, jagged, have a distortion effect, or have an unnatural clarity. 

\noindent \textbf{3. Inconsistent font}. The font of the tampered text may be slightly different in color, size, thickness, brightness, or style from the surrounding authentic text. 

\noindent \textbf{4. Inconsistent alignment}. The tampered text may have inconsistent spacing with the surrounding text or a small offset to the text line. 

\noindent \textbf{5. Text incoherence}. Tampered text may break the coherence of the sentence. 

\noindent \textbf{6. Lack of integration}. The tampered text may appear unnaturally placed and not integrated with its surroundings, or it may not blend seamlessly with its surroundings, appearing artificially overlaid or unnaturally pasted. 
Don't mention the image B in your answer, always assume that you are only observing the input image A.

As shown in Figure 1, our proposed prompt can help GPT4o output a satisfactory anomaly description.

\section{Detection Performance}
We present the detection performance of the detector in the proposed TextSleuth in Table~\ref{tab: det}. The precision, recall and F1-score under the ICDAR2017 DetEval protocol~\cite{nayef2017icdar2017} and an IoU threshold of 0.5 are used.

\begin{table}[h!]
\centering
\caption{The detection performance of the detector in the proposed TextSleuth. The IoU threshold is set to 0.5.}
\begin{tabular}{ccccccc}
\hline
Dataset   &  & Precision &  & Recall &  & F1-score \\ \cline{1-1} \cline{3-3} \cline{5-5} \cline{7-7} 
ETTD-Test &  & 0.994     &  & 0.986  &  & 0.990    \\
ETTD-CD   &  & 0.983     &  & 0.995  &  & 0.989    \\ \hline
\end{tabular}
\label{tab: det}
\end{table}

\section{Visualization}
\begin{figure*}[ht!]
 \centering
 \includegraphics[width=1.0\textwidth]{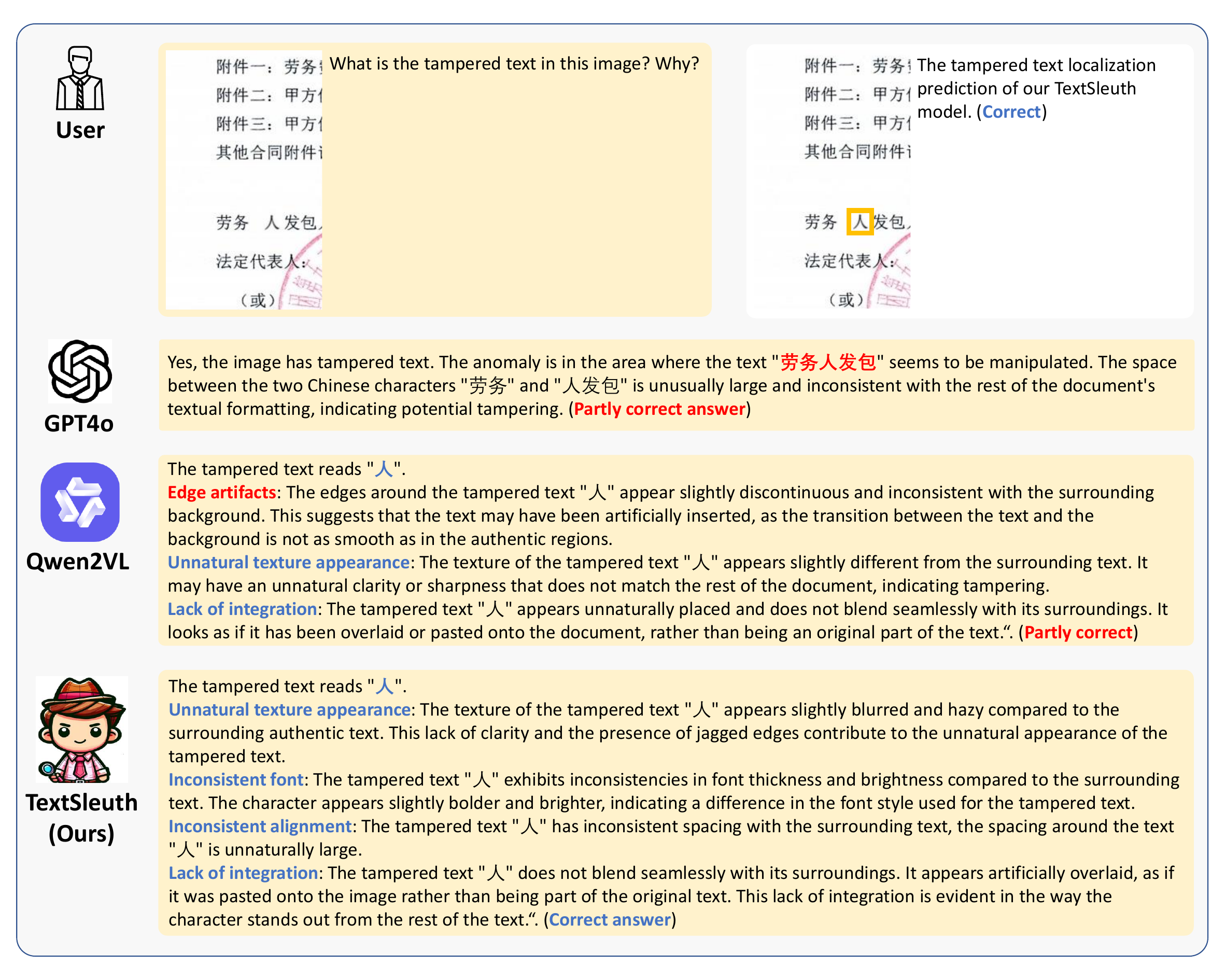}
 \caption{Our proposed TextSleuth can accurately detect the tampered text and better explain the anomalies of the tampered text.}
 \label{fig: comp1}
 \vspace{-0.3cm}
 \end{figure*}

\begin{figure*}[ht!]
 \centering
 \includegraphics[width=1.0\textwidth]{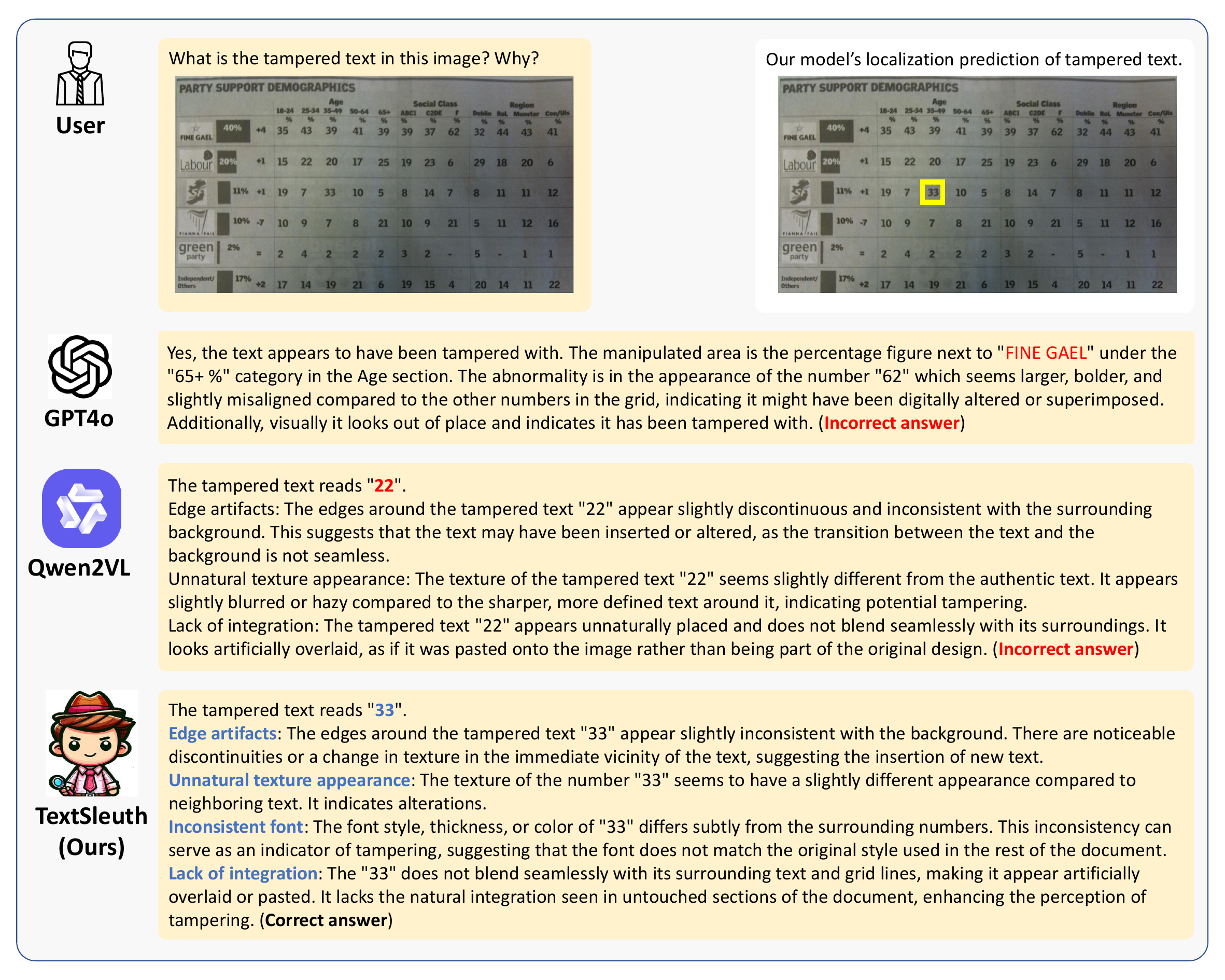}
 \caption{Our proposed TextSleuth can accurately detect the tampered text and better explain the anomalies of the tampered text.}
 \label{fig: comp2}
 \vspace{-0.3cm}
 \end{figure*}

\begin{figure*}[ht!]
 \centering
 \includegraphics[width=1.0\textwidth]{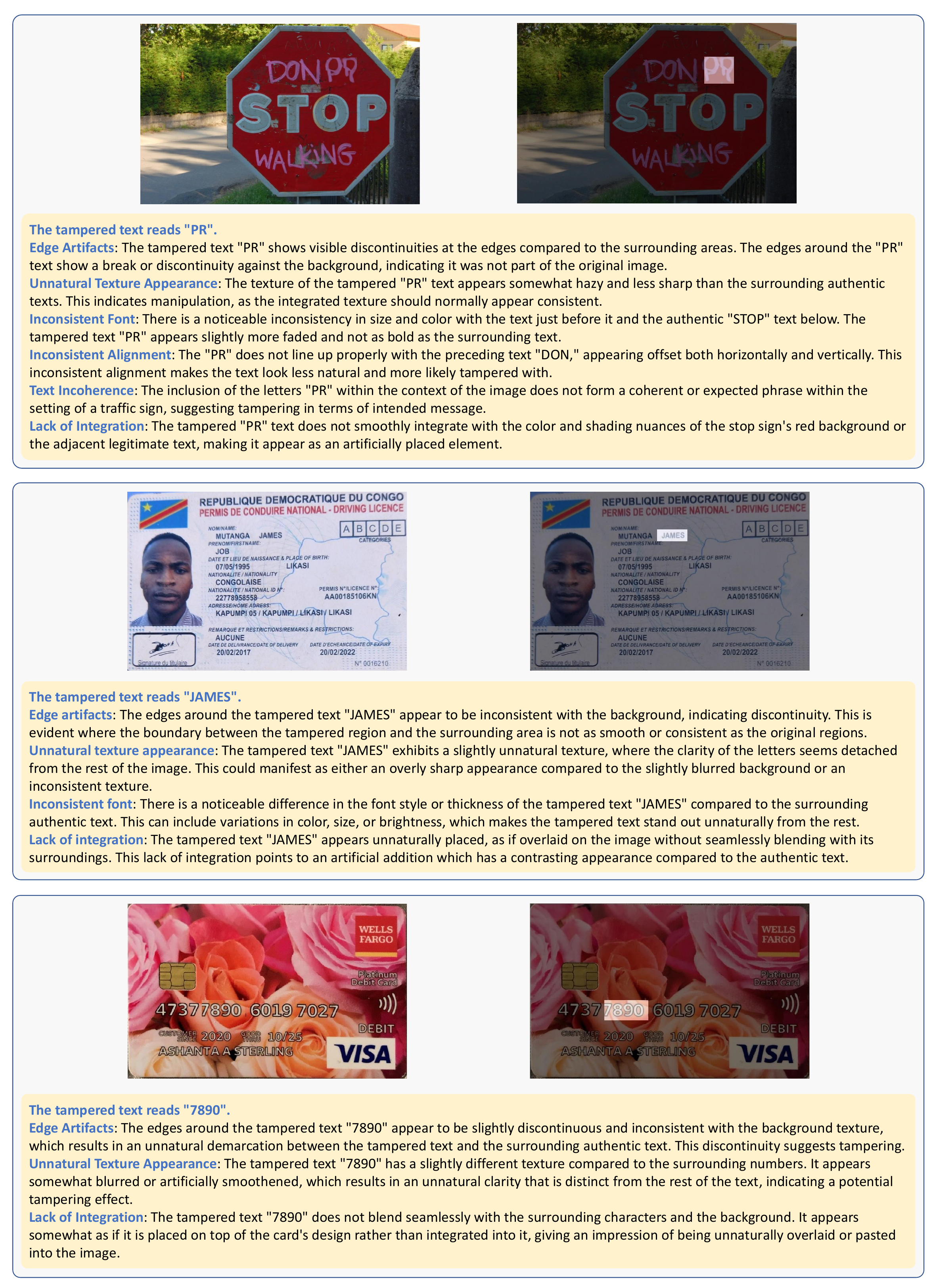}
 \caption{More data samples in the proposed ETTD dataset. The left image in each row is the original image and the tampered text region is highlighted in the right image.}
 \label{fig: data1}
 \vspace{-0.3cm}
 \end{figure*}

\begin{figure*}[ht!]
 \centering
 \includegraphics[width=1.0\textwidth]{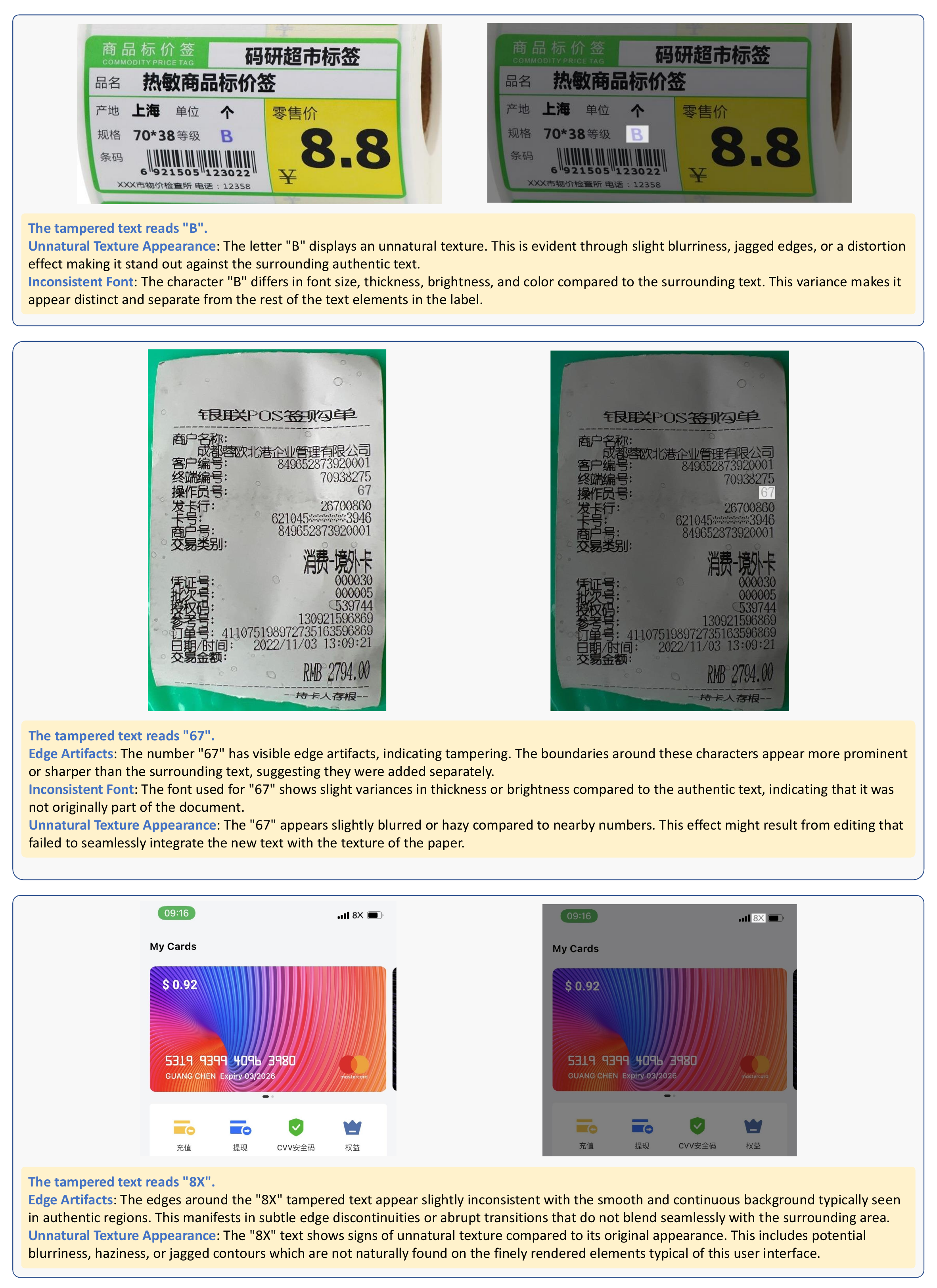}
 \caption{More data samples in the proposed ETTD dataset. The left image in each row is the original image and the tampered text region is highlighted in the right image.}
 \label{fig: data2}
 \vspace{-0.3cm}
 \end{figure*}

The prediction visualization of GPT4o, Qwen2VL-7B and our TextSleuth is shown in Figures~\ref{fig: comp1} and \ref{fig: comp2}. Evidently, the proposed method can produce more accurate results. More samples of the ETTD dataset are shown in Figures~\ref{fig: data1} and~\ref{fig: data2}.

\end{document}